%% file: root.tex
\documentclass[letterpaper, 10 pt, conference]{ieeeconf}  

\IEEEoverridecommandlockouts                              

\usepackage{framed,multirow}

\usepackage{amssymb}
\usepackage{caption}
\usepackage{latexsym}
\usepackage{bbm}
\usepackage{times}
\usepackage{epsfig}
\usepackage{graphicx}
\usepackage{amsmath}
\usepackage{subfigure}
\usepackage[english]{babel}
\usepackage{booktabs}
\usepackage{lineno}
\usepackage{url}
\usepackage{xcolor}
\usepackage{rotating}

\title{\LARGE \bf
ISA$^2$: Intelligent Speed Adaptation from Appearance}

\author{Carlos Herranz-Perdiguero$^{1}$ and Roberto J. L\'opez-Sastre$^{1}$
\thanks{*This work is supported by project PREPEATE, with reference TEC2016-80326-R, of the Spanish Ministry of Economy, Industry and Competitiveness. We gratefully acknowledge the support of NVIDIA Corporation with the donation of a GPU used for this research. Cloud computing resources were kindly provided through a Microsoft Azure for Research Award.}
\thanks{$^{1}$The authors are with GRAM research group, Department of Signal Theory and Communications,
         University of Alcal\'a, 28805, Alcal\'a de Henares, Spain
         {\tt\small c.herranz,@edu.uah.es, robertoj.lopez@.uah.es}}%
}

\def\eg{\emph{e.g. }}

\def\etal{\emph{et al. }}

\begin{document}

\maketitle
\thispagestyle{empty}
\pagestyle{empty}

\input{abstract}

\input{introduction}
\input{relatedwork}
\input{dataset}
\input{approach}
\input{experiments}
\input{conclusion}

{\small
\bibliographystyle{IEEEtran}
\bibliography{IEEEexample}
}

\end{document}

%% file: abstract.tex
\begin{abstract}
In this work we introduce a new problem named Intelligent Speed Adaptation from Appearance (ISA$^2$). Technically, the goal of an ISA$^2$ model is to predict for a given image of a driving scenario the \emph{proper} speed of the vehicle. Note this problem is different from predicting the actual speed of the vehicle. It defines a novel regression problem where the appearance information has to be directly mapped to get a prediction for the speed at which the vehicle should go, taking into account the traffic situation. First, we release a novel dataset for the new problem, where multiple driving video sequences, with the annotated adequate speed per frame, are provided. We then introduce two deep learning based ISA$^2$ models, which are trained to perform the final regression of the proper speed given a test image. We end with a thorough experimental validation where the results show the level of difficulty of the proposed task. The dataset and the proposed models will all be made publicly available to encourage much needed further research on this problem.
\end{abstract}

%% file: introduction.tex
\section{INTRODUCTION}
\label{sec:intro}

For years, speed has been recognized as one of the three main contributing factors to deaths on our roads. In fact, 72 \% of road traffic accidents in the city could be prevented with an adequate vehicle speed, according to the MAPFRE Foundation \cite{mapfre}. Furthermore, the European Transport Safety Council (ETSC) claims that speed is the cause of the death of 500 people every week on European roads \cite{etsc-report}. So, to control the speed of our vehicles, using an Intelligent Speed Adaptation (ISA) system, should be a high-priority research line.

A research by the Norwegian Institute for Transport Economics \cite{Vaa2014} advocates the benefits of an ISA system, which the study found to be the most effective solution in saving lives. Some studies of the ETSC reveal that the adoption of the ISA technology is expected to reduce collisions by 30\% and deaths by 20\% \cite{etsc-study}.

Off-the-shelf ISA solutions use a speed traffic sign recognition module, and/or GPS-linked speed limit data to inform the drivers of the current speed limit of the road or highway. However, these solutions have the following limitations. First, GPS information is inaccurate and may not be correctly updated. For example, an ISA model based only on GPS information would have difficulties in certain urban scenes with poor satellite visibility, or in distinguishing whether the vehicle is in a highway lane or on the nearby service road, where the speed limit has to be drastically reduced. It is true that a speed traffic sign recognition module can mitigate some of these problems, but for doing so we need to guarantee the visibility of the signs. Second, they provide only the speed limit of the road, but not the speed \emph{appropriate} to the actual traffic situation.

\begin{figure}[]
	\includegraphics[width=\linewidth]{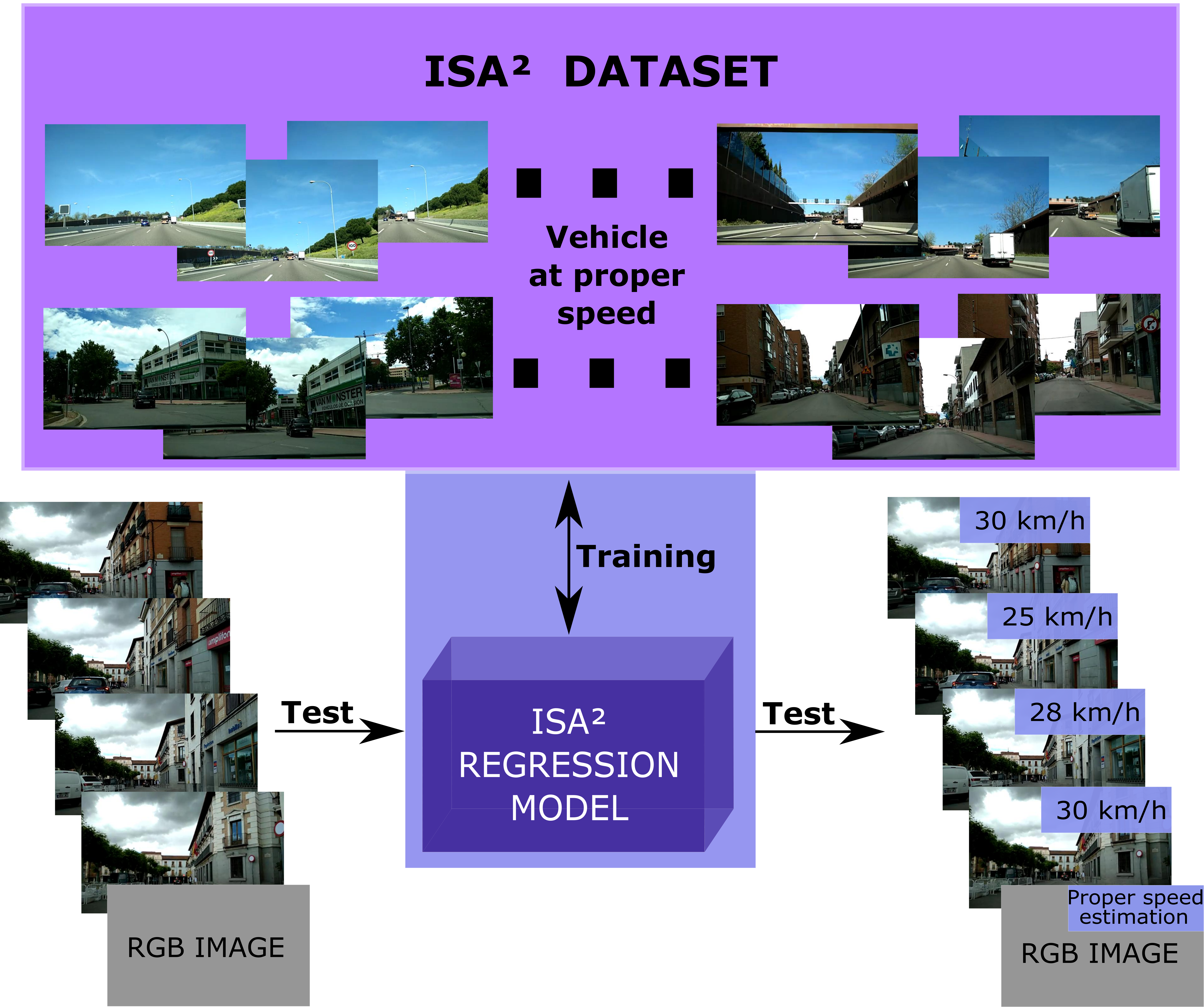}
	\caption{ISA$^2$ problem. An ISA$^2$ model must be able to perform a regression of the \emph{adequate} speed of the vehicle, inferring it just using the appearance information of the image. It has to be trained on video sequences providing the proper speed for the traffic situation, to be able to provide an estimation of the adequate speed on test images.}
	\label{fig:abstract}
\end{figure} 

To address all these limitations, in this paper we propose a new paradigm for the ISA models, called \textbf{ISA} from \textbf{A}ppearance, or ISA$^2$. Technically, as it is shown in Figure \ref{fig:abstract}, we introduce the idea of learning a regression function able to map the images to a speed adequate to the traffic situation. For doing so, we need to train and evaluate the ISA$^2$ solutions using a dataset with video sequences that show a driving behaviour that is appropriate to the real traffic situation. The proposed problem is actually very challenging. Could a human, from a single image, discern between whether a vehicle should go at 80 or 110 km/h on a motorway according to the actual traffic? 

The main contributions of our work are as follows:
\begin{enumerate}
 \item To the best of our knowledge, we propose for the first time the novel problem of inferring the \emph{adequate} speed of a vehicle from just an image.
 \item We introduce two deep learning based ISA$^2$ models, which are trained to perform the final regression of the proper speed for the vehicle. One consists in learning a deep network to directly perform the speed regression. The other approach is based on a deep learning model to obtain a semantic segmentation of the traffic scene. We then combine this output with a spatial pyramid pooling strategy to build the features used to learn the regressor for the proper speed.
 \item We also release a novel dataset for the new ISA$^2$ problem, where the proposed models are evaluated. We conduct an extensive set of experiments and show that our ISA$^2$ solutions can report an error for the prediction of the speed lower than 6 km/h.
\end{enumerate}

The rest of the paper is organized as follows. In Section \ref{sec:relatedwork}, we discuss related work. In Section \ref{sec:dataset} we describe the ISA$^2$ dataset and the evaluation protocol. Our ISA$^2$ models are detailed in Section \ref{sec:approach}. We evaluate our models, and analyze their performance in Section \ref{sec:exp}. We conclude in Section \ref{sec:conclusion}.

%% file: relatedwork.tex
\begin{figure*}[t]
	\centering
	\captionsetup[subfigure]{justification=centering}
	\subfigure[Highway]{\label{fig:highway}\includegraphics[width=0.48\textwidth] {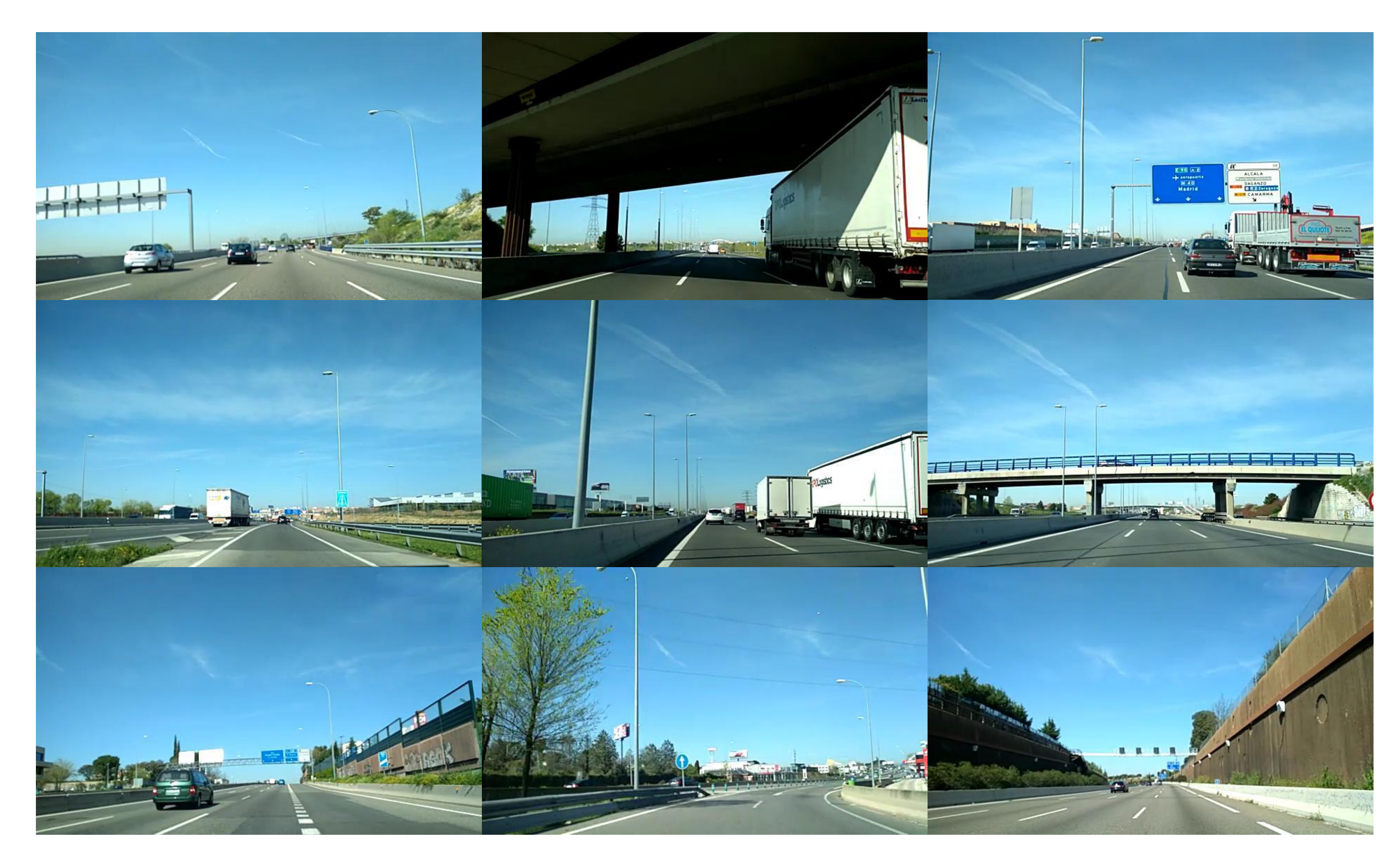}}
	\subfigure[Urban]{\label{fig:urban}\includegraphics[width=0.48\textwidth] {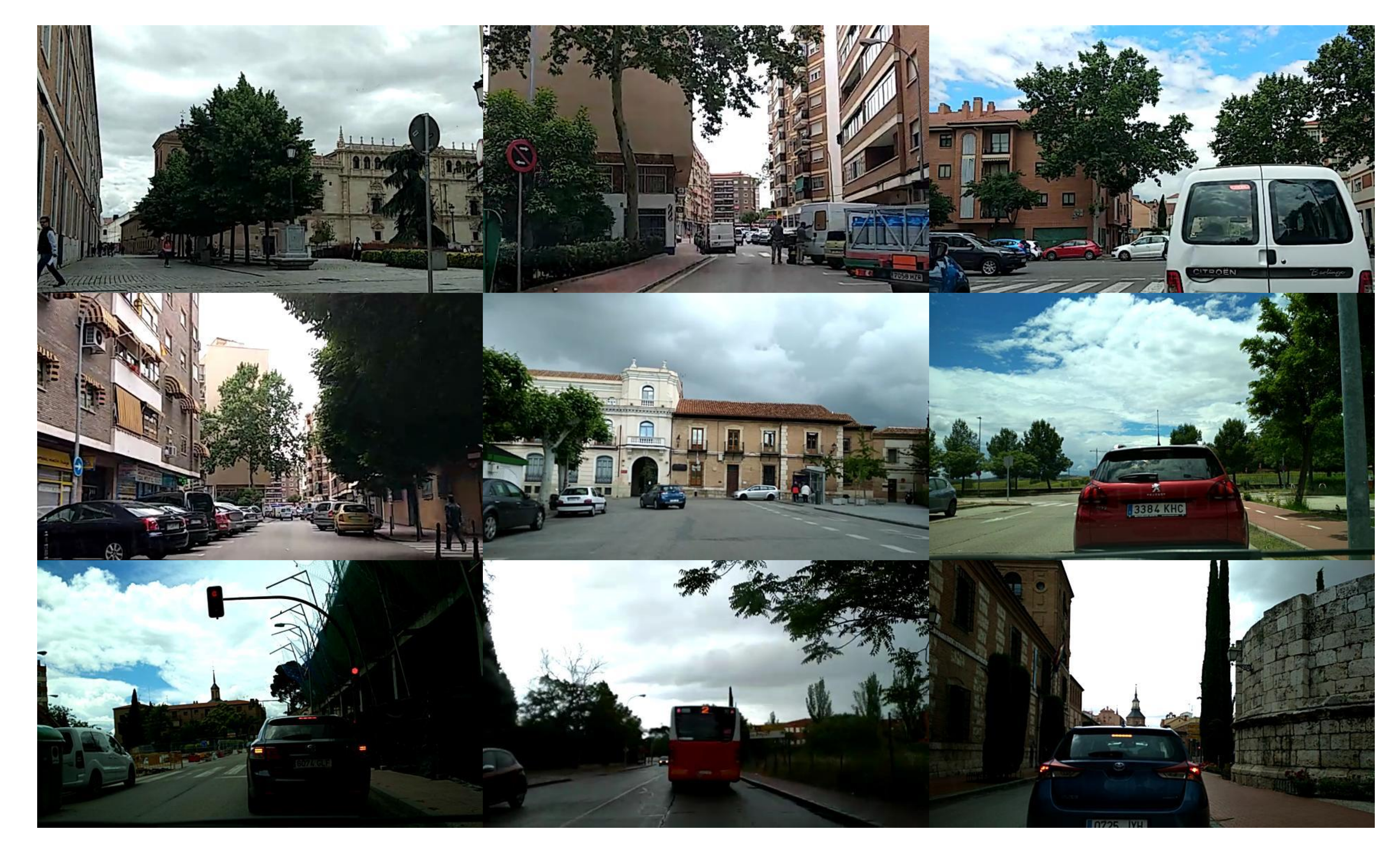}}
	\caption{Set of images from the ISA$^2$ dataset in highway and urban environments.}
	\label{fig:ISA2_images}
\end{figure*}
\section{RELATED WORK}
\label{sec:relatedwork}

Although being able to estimate the appropriate speed for a vehicle is a key task for the automotive industry, which year after year is increasing the budget for R\&D projects in its pursuit to achieve a fully autonomous vehicle, there are no previous works that seek to predict this speed just using images or visual information.

In the literature, we can find some works that deal with the different problem of learning a generic driving model, \eg \cite{Sierra-Gonzalez2018,Xu2017,Codevilla2018}.

Probably, the closest works we can find to the problem we are trying to solve, focus on estimating the \emph{actual} speed of a vehicle, which is a different problem anyhow. Several techniques have been proposed for this purpose, from the design of image processing methods using optical flow \cite{Chhaniyara2008,Shukla2012,Sreedevi2011} to proposals for motion estimation based on the subtraction of the background \cite{Atkociunas2005}. Chhaniyara \etal \cite{Chhaniyara2008} focus on robotics platforms moving over different types of homogeneous terrains such as fine sand, coarse sand, gravel, etc. The rest of works \cite{Shukla2012,Sreedevi2011,Atkociunas2005} have been designed to estimate the speed of vehicles from video sequences acquired with a fixed video surveillance camera.

We, instead, propose to estimate the \emph{proper} speed for a vehicle, according to the traffic situation, by using a vehicle \emph{on-board} camera. While all the works mentioned above aim to estimate the actual speed at which the vehicle is moving, our ISA$^2$ models need to estimate the appropriate speed at which the vehicle should go. Our goal is not to know how fast a car goes, but how fast it should go.

%% file: dataset.tex
\section{ISA$^2$ DATASET}
\label{sec:dataset}

Here, we introduce the novel ISA$^2$ dataset, which allows us to train and test different approaches for the new challenging ISA$^2$ problem.

The database consists of 5 video sequences taken from both urban and interurban scenarios in the Community of Madrid, Spain. In total, we provide a set of 149.055 frames, with a size of $640\times384$ pixels, with the annotation of the proper speed of the car (km/h). During the driving for the acquisition of the dataset, in addition to respecting the speed limits, our driver has carefully tried to adjust the speed of the vehicle to what he considers to be an appropriate speed, according to the traffic situation. Figures \ref{fig:highway} and \ref{fig:urban} show some images of both, highway and urban routes, respectively.

To structure the database, both scenarios have been split into training and test subsets. For the 3 urban recordings, we use two of them for training/validation, an the third one for testing. We also provide two highway recordings, one for training/validation and the other for testing. These splits between training and testing have been done so that different scenarios and circumstances are well represented in both sets. Those scenarios include maximum and minimum speed over the sequences, stops at traffic lights or entrances and exists on the highway using service roads, for instance. Finally, with the aim of evaluating how well the different approaches are able to generalize, we introduce unique factors in the test subsets, such as, different weather conditions (rain) in the urban test set. All these aspects clearly help to release a challenging dataset. Table \ref{tab:speed_routes} shows the mean speed of the vehicle for the different subsets described.

\begin{table}[h]
	\centering
	\vspace*{0.25cm}
	\setlength{\tabcolsep}{10pt}
	\renewcommand{\arraystretch}{1.3}
	\caption{Mean speed and standard deviation of the different sets in the $ISA^2$ dataset}
	\label{tab:speed_routes}
	\scalebox{0.8}{
	\begin{tabular}{  c*{3}{ c }}	
		\toprule
		\textbf{Route} & \textbf{Set} & \textbf{Mean speed (km/h)} & \textbf{Std. deviation (km/h)} \\ 
		\midrule
		Highway & Training & 84.31 & 18.15 \\
		\midrule 
		Highway & Test & 95.08 & 12.81\\
		\midrule 
		Urban & Training & 19.55 & 13.60 \\
		\midrule 
		Urban & Test & 19.59 & 14.78\\
		\bottomrule 		
	\end{tabular}
	}
\end{table}

%% file: approach.tex
\section{MODELS FOR ISA$^2$}
\label{sec:approach}

Our main objective during the design of the ISA$^2$ models is to propose a strong visual representation that allows the models to predict the appropriate speed for the vehicle. 

The ISA$^2$ problem starts with a training set of images $S = \{(I_{i}, s_i)\}_{i = 1}^{N}$, where $N$ is the number of training samples. For each sample $i$ in the dataset, $I_i$ represents the input image, and $s_{i} \in \mathbb{R}$ encodes the annotation for the speed.

\begin{figure}[t]
	\includegraphics[width=\linewidth]{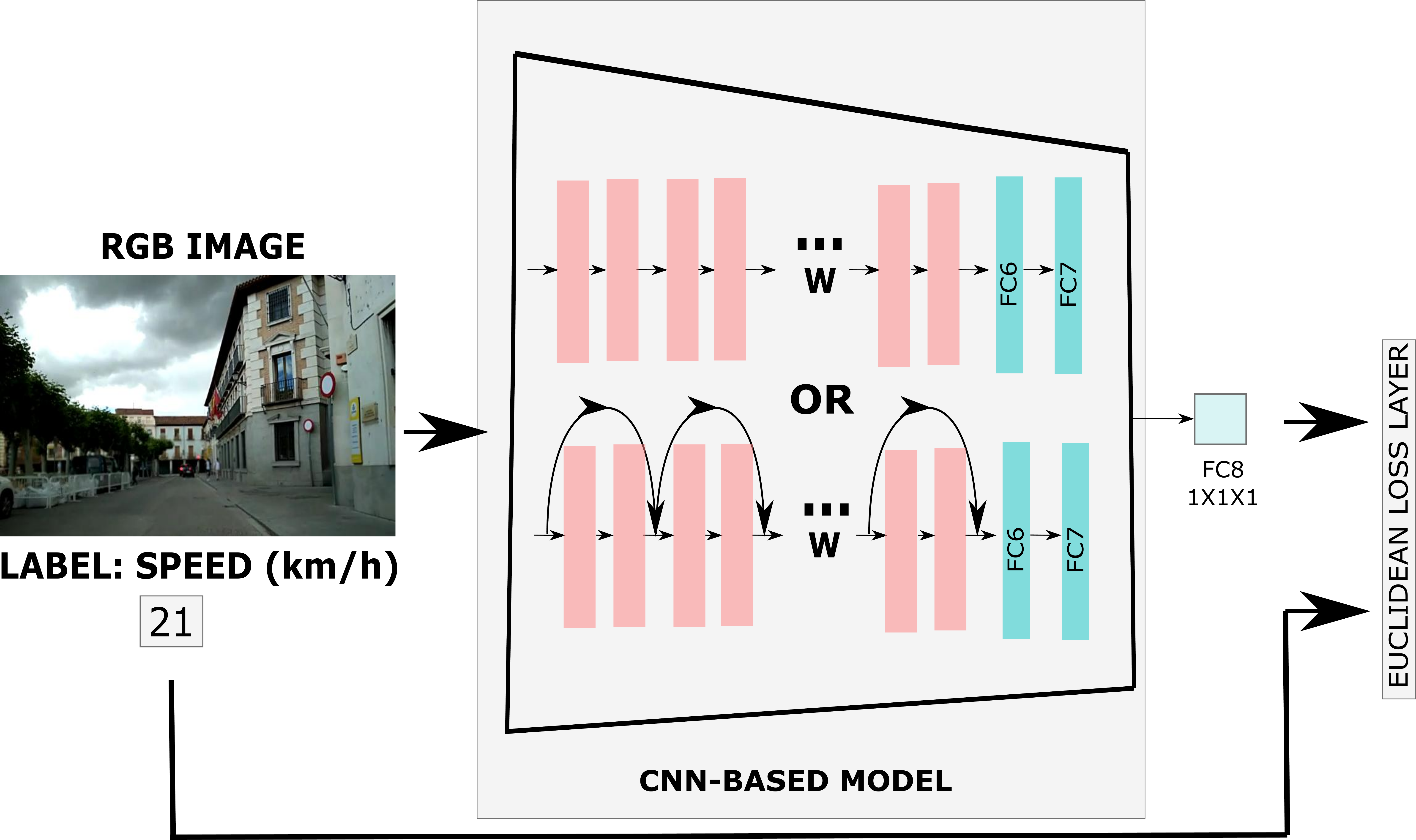}
	\caption{ISA$^2$ from a CNN based architecture for regression.}
	\label{fig:ISA2_CNNS}
\end{figure} 

We first propose to learn a Convolutional Neural Network (CNN) \cite{LeCun1990} to directly perform the regression of the adequate speed. Technically, as it is shown in Figure \ref{fig:ISA2_CNNS}, we use two different architectures: a VGG-16 \cite{Simonyan2015} or a Residual CNN \cite{He2016} (ResNet). Therefore, our networks are trained to learn the direct mapping from the image to the speed $\hat{s}$, a function that can be expressed as follows,

\begin{equation}
\hat{s}_{W} = f(W,I_i) \, ,
\label{eq:prediction}
\end{equation}
where, $f(W,I_i): I_i \rightarrow \mathbb{R}$ represents the mapping that the network performs to the input images. We encode in $W$ the trainable weights of the deep architecture. We replace the loss function of the original network designs, which is no longer a softmax, but a loss based on the Euclidean distance.

\begin{figure}[t]
	\includegraphics[width=\linewidth]{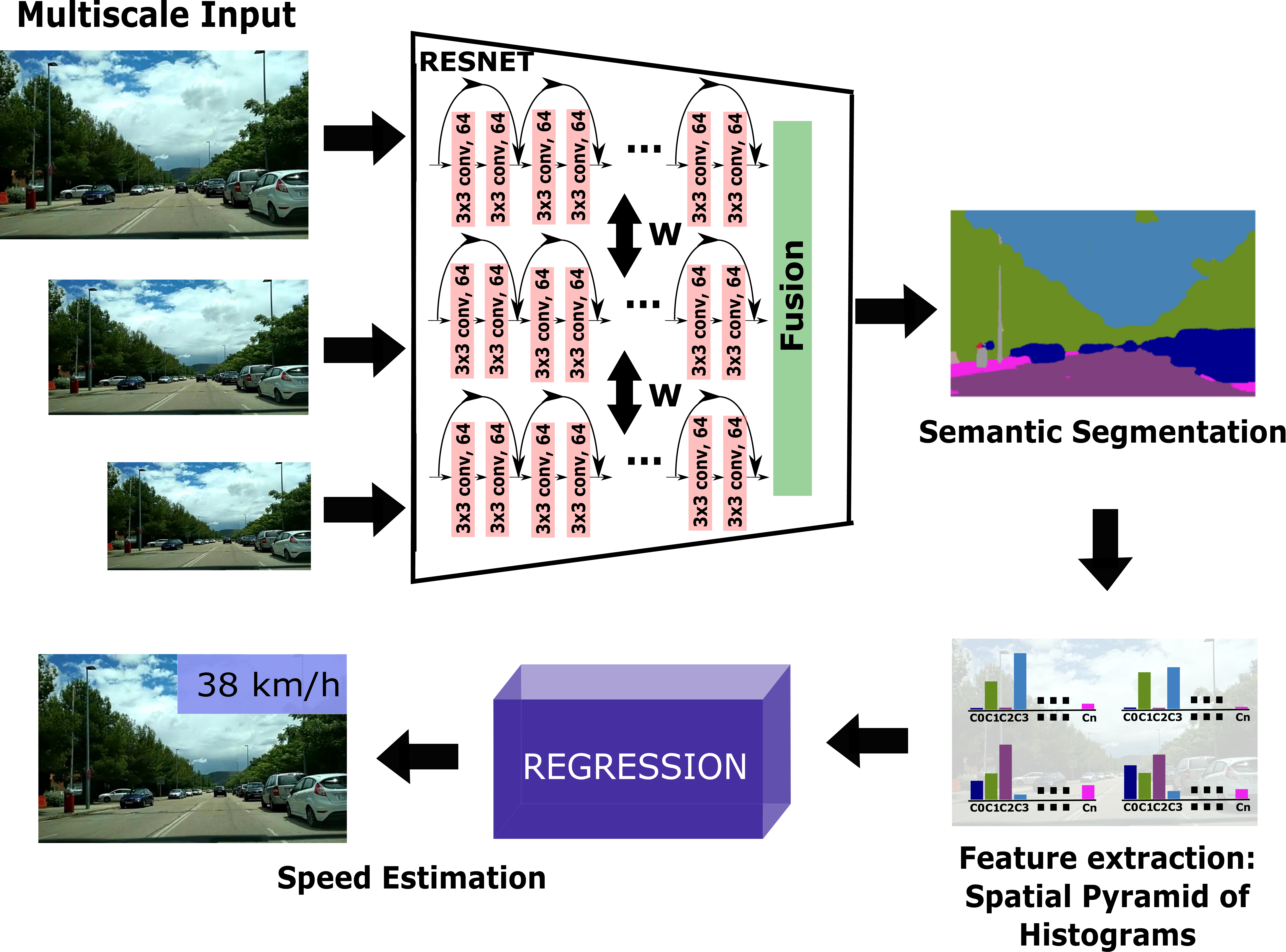}
	\caption{ISA$^2$ from a semantic segmentation and a regressor.}
	\label{fig:ISA2_Histograms}
\end{figure} 

The second approach is mainly based on a semantic segmentation model, see Figure \ref{fig:ISA2_Histograms}. Our system starts performing a dense pixel labeling of the traffic scene. We then use a spatial pyramid pooling strategy, to build a descriptor for the image, which is based on the histogram of the different labels produced by our semantic segmentation model. This descriptor is used to learn a final regressor, which is the one in charge of the prediction of the proper speed.

Technically, for this second approach, we first implement the DeepLab \cite{He2016} model, using a ResNet-101 as the base network. We train the DeepLab using a multi-scale input, using the scale factors $\{0.5, 0.75, 1\}$. We then fuse the prediction for each scale, taking the maximum response given by the network for each scale. Note that the ISA$^2$ dataset does not provide semantic segmentation annotations, therefore this model is trained using the Cityscapes dataset \cite{cityscapes}.

For the final regression, we evaluate in the experiments several approaches: linear regressor, lasso regressor, boosting trees and linear Support Vector Regressors. For all of them, we evaluate the impact of adding spatial information by using spatial pyramid pooling of up to 3 levels.

%% file: experiments.tex
\section{EXPERIMENTS}
\label{sec:exp}
To evaluate the effectiveness of our models, we use here the ISA$^2$ dataset. We detail the experimental setup and main results in the following sections.

\subsection{Experimental setup and evaluation metric}

For our CNN-based approaches, VGG and ResNet-101, we fine-tune pre-trained models on the large-scale ImageNet dataset \cite{imagenet}. Both networks are trained for 4K iterations with a learning rate of $10^{-4}$ for the first 2K iterations, and  of $10^{-5}$ for the rest. We use stochastic gradient descent (SGD) with a momentum of 0.9 and a batch size of 20 images for both architectures.

With respect to our models based on the semantic segmentation of the images, we cross validate both the specific parameters of the different regression methods and the spatial pyramid levels we use.

To measure the performance of the different models, we use the standard Mean Absolute Error (MAE) metric, which is defined as the difference in absolute value between the real speed, $s_r$, and the proper speed estimated by an ISA$^2$ model, $\hat{s}$, averaged for the $K$ images of the test set, according to:

\begin{equation}
\frac{1}{K}\sum_{i=1}^{K}|s_{r_i} - \hat{s}_i| .
\end{equation}

We evaluate the MAE independently for the urban and highway set of images, because this provides a more detailed analysis of the results. 

\subsection{Quantitative results}

In Table \ref{tab:joint_results} we present the results of our ISA$^2$ approaches. In general, we show that our second approach, that is a semantic segmentation (SS) plus a regressor, obtains better results, only for the urban scenarios, than the first model proposed, where the CNNs directly cast the speed estimation. In a highway setting, our first approach reports a lower MAE. Probably, the fact that our first type of approaches have more parameters, allows them to adjust better the prediction to both types of environments.

\begin{table}[h]
	\centering
	\vspace*{0.25cm}
	\setlength{\tabcolsep}{15pt}
	\renewcommand{\arraystretch}{1.35}
	\caption{MAE comparison of our different proposed methods. For each model, we train a unique regressor for both highway and urban scenarios.}
	\label{tab:joint_results}
	\scalebox{0.9}{
	\begin{tabular}{  l*{3}{ c }}	
		\toprule
		\textbf{\multirow{2}{*}{Method}} & \textbf{Urban MAE} & \textbf{Highway MAE} \\
		& (Km/h) &  (Km/h) \\ \hline
		VGG-16 & 12.58 & \textbf{11.57} \\		 
		ResNet-101 & 11.49 & 11.87 \\
		\midrule
		SS + Linear regression & 9.15 & 15.78 \\
		 
		SS + SVR & 10.69 & 16.76 \\
		 	
		SS + Lasso regression & \textbf{8.74} & 18.13 \\
		 	
		SS + Boosting Trees & 9.78 & 13.86\\ 
		\bottomrule
	\end{tabular}
	}
\end{table}

\begin{figure*}
	\centering
	\captionsetup[subfigure]{justification=centering}
	\subfigure[Joint training]{\includegraphics[width=0.48\linewidth, trim={3cm 9cm 3cm 9cm}, clip]{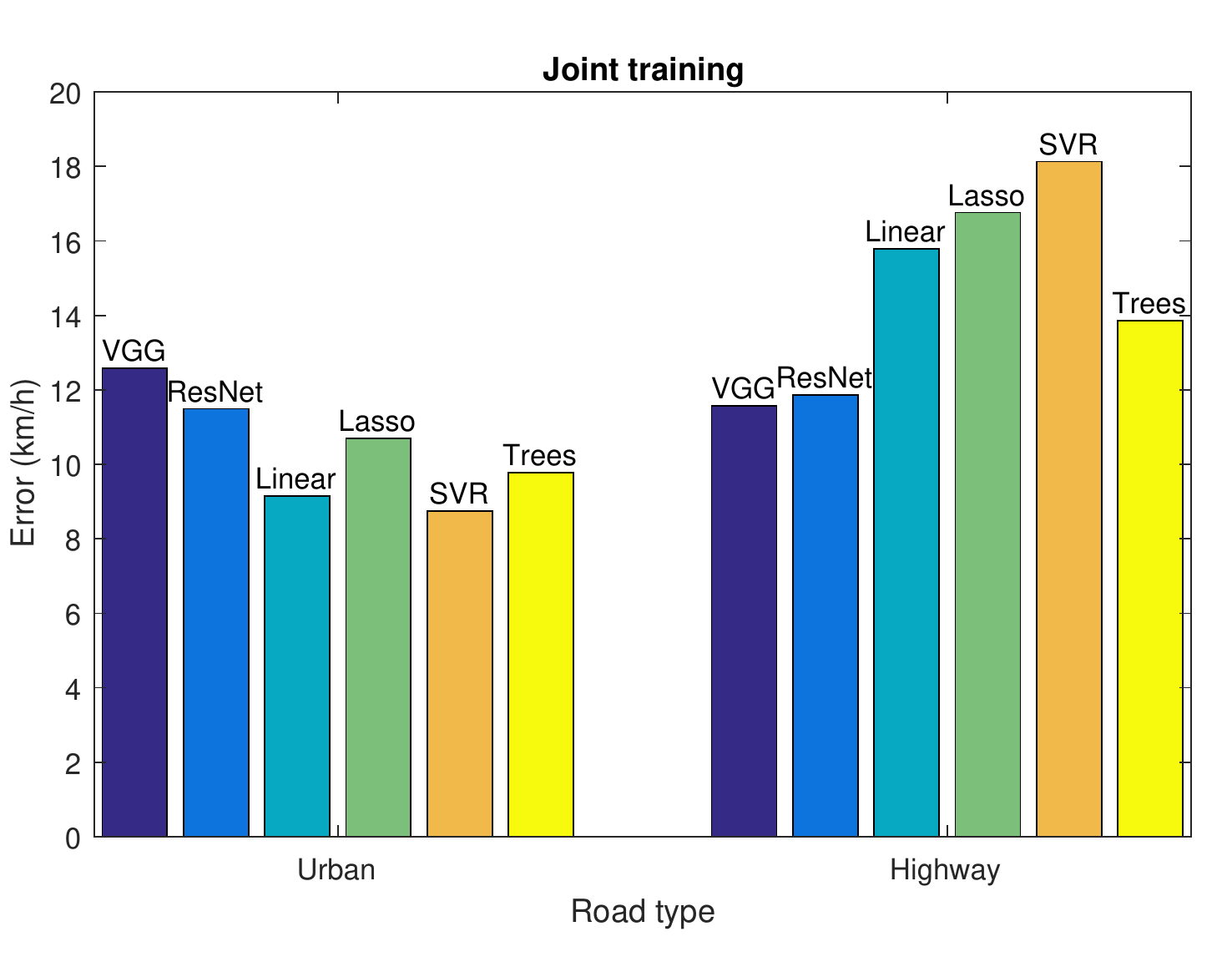}}
	\subfigure[Independent training]{\includegraphics[width=0.48\linewidth, trim={3cm 9cm 3cm 9cm}, clip]{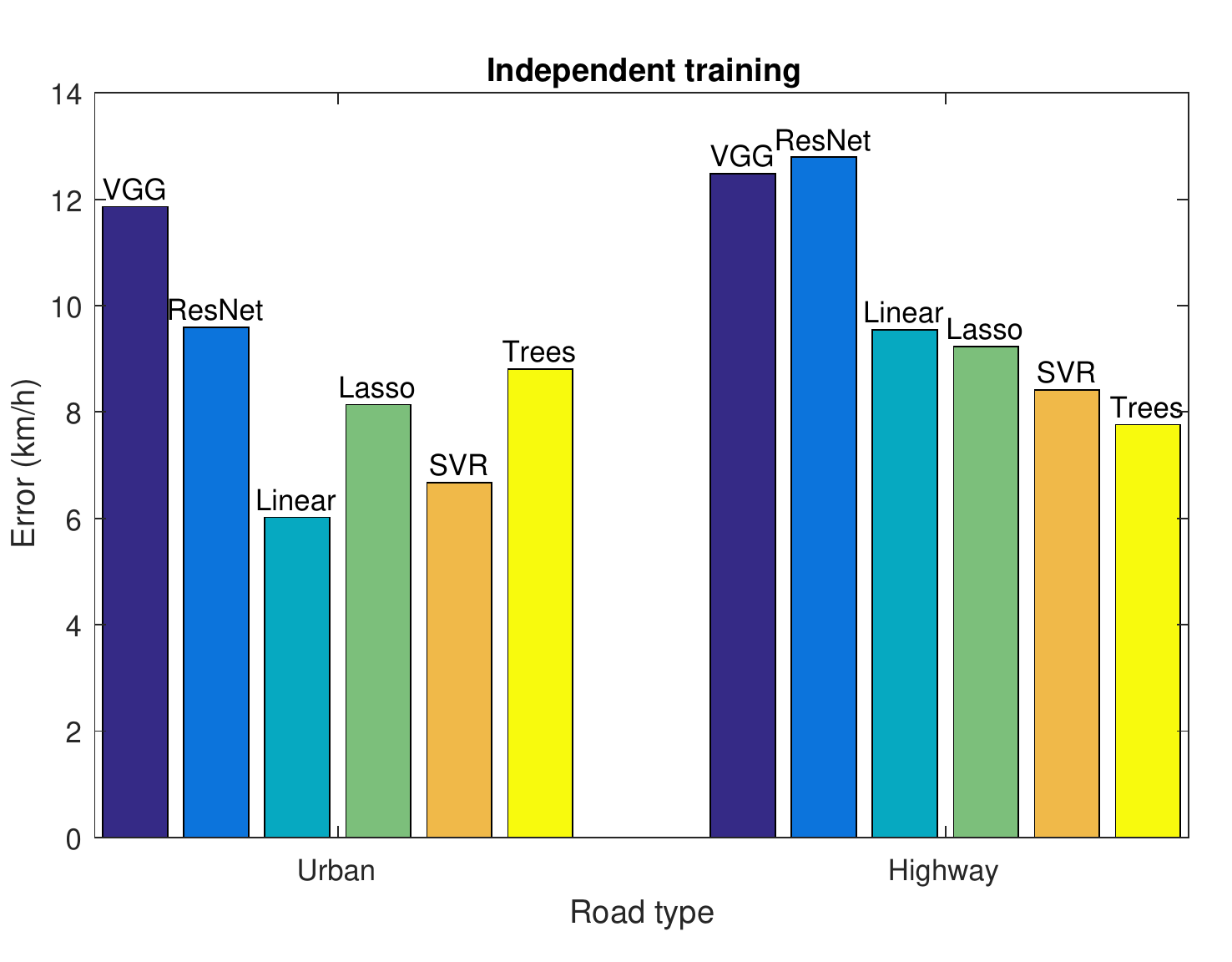}}
	\caption{MAE comparison between all of our different approaches to the ISA$^2$ problem.}
	\label{fig:graph_results}
\end{figure*}

In this sense, we decide to perform a second experiment. We proceed to train an ISA$^2$ model for each type of scenario (urban and highway) separately. Table \ref{tab:independant_results} shows the results. Now, models based on the SS perform better for both urban and highway images. In highway images, boosting trees are the ones that offer the best results, followed by the lasso regression and the SVR. On the other hand, in the urban sequences, a linear regression exhibits the best performance, followed by the lasso regression and the SVR. As a conclusion, it is clear that for our models based on SS, it is beneficial to train a regressor for each type of scenario separately. Figure \ref{fig:graph_results} shows a graphical comparison of the results, following the two training methods described.

\begin{table}[h]
	\centering
	\vspace*{0.25cm}
	\setlength{\tabcolsep}{15pt}
	\renewcommand{\arraystretch}{1.35}
	\caption{MAE comparison of our different proposed methods. For each model, we train an independent regressor for highway and urban scenarios.}
	\label{tab:independant_results}
	\scalebox{0.9}{
	\begin{tabular}{  l*{3}{ c }}	
		\toprule
		\textbf{\multirow{2}{*}{Method}} & \textbf{Urban MAE} & \textbf{Highway MAE} \\
		& (Km/h) &  (Km/h) \\ \hline
		VGG-16 & 11.86 & 12.48 \\
		 
		ResNet-101 & 9.59 & 12.79 \\
		\midrule 
		SS + Linear regression & \textbf{6.02} & 9.54 \\
		 
		SS + SVR & 8.14 & 9.23\\ 
		 	
		SS + Lasso regression & 6.67 & 8.72\\ 
		 	
		SS + Boosting Trees & 8.81 & \textbf{7.76}\\ 
		\bottomrule
	\end{tabular}
	}
\end{table}

\begin{figure*}
	\includegraphics[width=\linewidth]{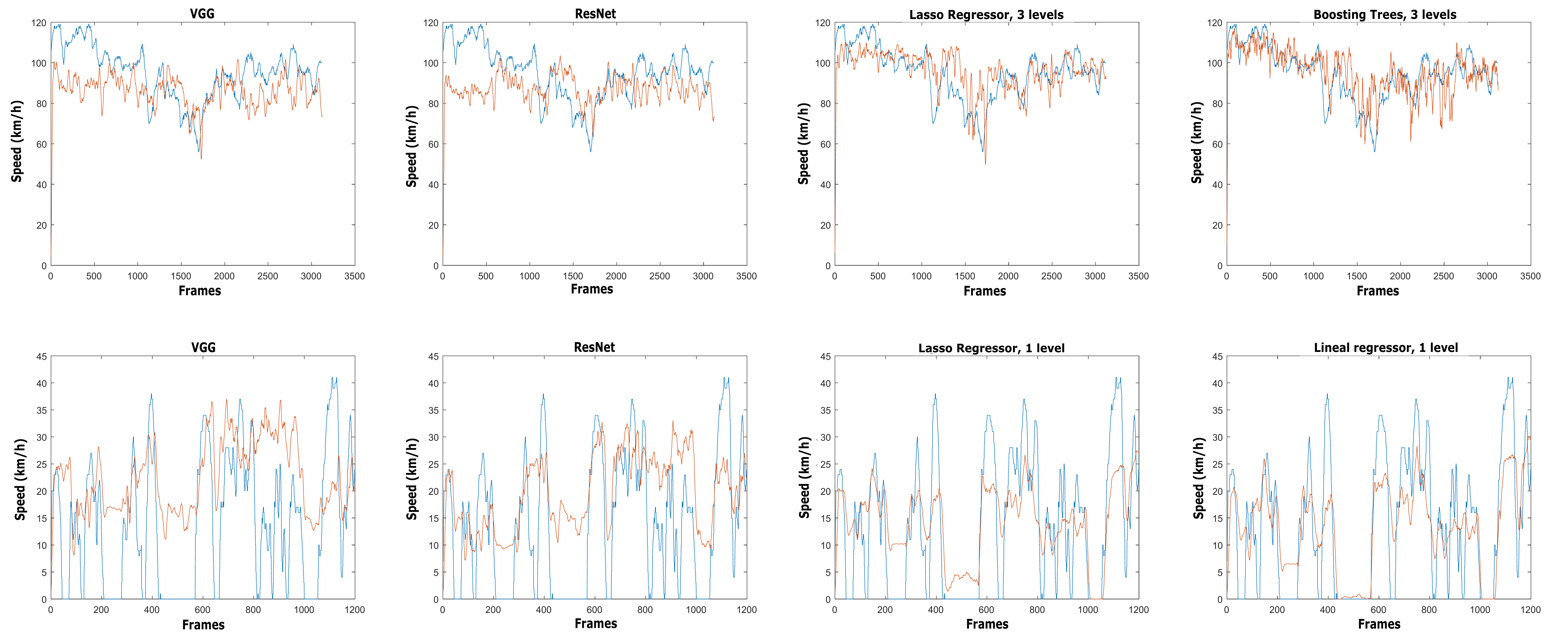}
	\caption{Proper speed (blue) vs. Estimated proper speed (red) of different methods. First row corresponds to the highway test sequence, while second row shows results on the urban sequence.}
	\label{fig:speed_results}
\end{figure*}

Finally, Figure \ref{fig:speed_results} shows a graphical comparison between the proper speed of the vehicle (in blue) and the estimated speed (in red) by the different ISA$^2$ models proposed. For each type of scenario, results of the two CNN-based models used are shown together with the two best models based on SS + regression.

Interestingly, for the highway test sequence, all our models detect that it is necessary to reduce the speed halfway along the route, at a time when the driver leaves the highway towards a service road, to finally rejoin a different highway. In general, we can observe that the neural networks have more difficultly to predict the proper speed, than the SS based solutions. This is particularly evident in the initial section of the routes, where the error made by the CNNs exceeds 30 km/h.

For the urban test sequence, it is remarkable that the CNNs are not capable of reducing the estimated proper speed when the vehicle is completely stopped, mainly at red traffic lights. On the other hand, SS-based regressors do adjust such situations much better.

\subsection{Qualitative results}

\begin{figure*}
	\includegraphics[width=\linewidth]{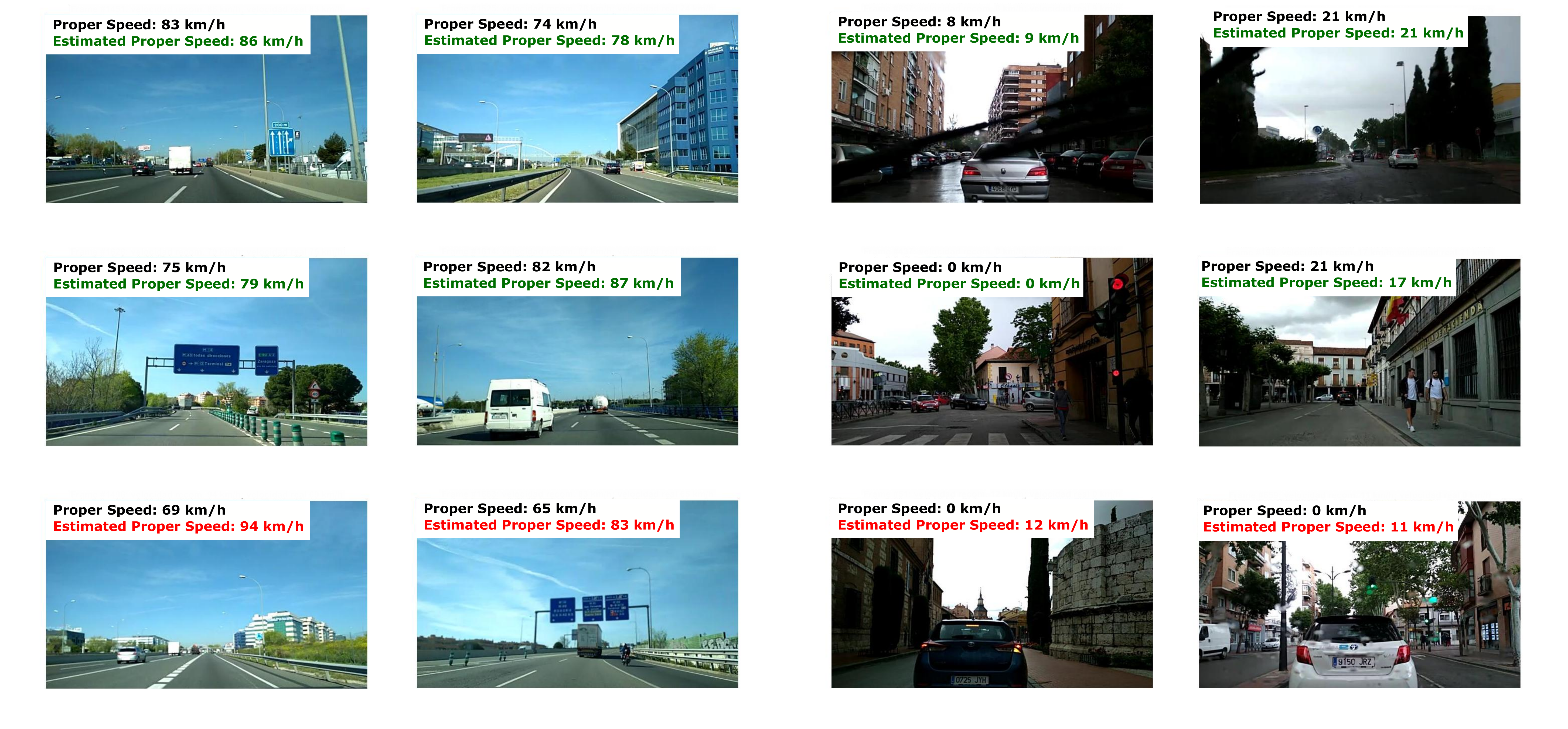}
	\caption{Qualitative results of our best models for both type of roads. First two rows show a set of frames for which our ISA$^2$ solutions obtain the best predictions. Last row shows moments in which the speed difference is high.}
	\label{fig:qual_results}
\end{figure*}

We show a set of qualitative results in Figure \ref{fig:qual_results}. Those results correspond to the best of our models for each type of road, i.e. using boosting trees in highway and SS + Linear regression in an urban environment.

Analyzing these results, we observe some of the difficulties our models have. On highways, for instance, the biggest errors for the estimation of the proper speed occur when the vehicle wants to leave the motorway, which leads the driver to slow down. Obviously, our models, which are based exclusively on what \textit{the vehicle sees} at any given time, are not able to anticipate the driver's intentions, so they estimate a speed higher than the real one. However, as soon as the driver leaves the motorway and change the type of road, the models do correctly adjust the speed.

In urban environments, the main problem is related to the presence of stationary vehicles on the road, which implies that our vehicle has to stop when it reaches them. In those cases, although there is a decrease in the estimated proper speed, the models do not come to realize that it is necessary to completely stop. This does not occur in the presence of red traffic lights, where the estimated proper speed reaches 0 km/h.

%% file: conclusion.tex
\section{CONCLUSION}
\label{sec:conclusion}

In this paper we propose for the first time the ISA$^2$ problem. It is a difficult and interesting problem, that has not been studied before. We also release a new dataset and propose an evaluation protocol to assist the research on ISA$^2$. Finally, we have introduced and evaluated two types ISA$^2$ models, and the results show the level of difficulty of the proposed task.

The dataset and the proposed models will all be made publicly available to encourage much needed further research on this problem.